\documentclass[letterpaper, 10 pt, conference]{ieeeconf}  % Comment this line out if you need a4paper

\IEEEoverridecommandlockouts                              % This command is only needed if 
\usepackage{times}
 \usepackage{booktabs}
 \usepackage{wrapfig}
 \usepackage{amsfonts}       
\usepackage{nicefrac}       
\usepackage{microtype}     
\usepackage{amsmath}
\usepackage{graphicx}
\usepackage{amssymb}
\let\labelindent\relax

\usepackage{enumitem}
\usepackage[inkscapeformat=png]{svg}
\definecolor{ms_red}{RGB}{167, 22, 34}
\definecolor{activity-color}{RGB}{188, 75, 81}
\definecolor{url-color}{RGB}{181, 101, 118}

\def\shownotes{1}  %set 1 to show author notes
\ifnum\shownotes=1
\newcommand{\authnote}[2]{{$\ll$\textsf{\footnotesize #1 notes: #2}$\gg$}}
\else
\newcommand{\authnote}[2]{}
\fi

\usepackage{hyperref}
\hypersetup{
    colorlinks=true,
    linkcolor=blue,
    urlcolor=url-color,
    }

\usepackage{lib}

\usepackage[numbers,sort]{natbib}

\newcommand{\algoName}{\text{\texttt{HOPMan}~}}
\newcommand{\wG}{\texttt{MG}}
\newcommand{\sG}{\texttt{G}}
\newcommand{\sTG}{\texttt{SG}}
\newcommand{\objectcategory}{\text{object category}}
\newcommand{\instantiation}{\text{object instance}}
\newcommand{\skill}{\text{skill}}

\title{\LARGE \bf
Towards Generalizable Zero-Shot Manipulation \\ via Translating Human Interaction Plans
}

\author{Homanga Bharadhwaj$^{1}$, Abhinav Gupta$^{*,2}$, Vikash Kumar$^{*2}$, Shubham Tulsiani$^{*,2}$% <-this % stops a space
\thanks{*equal contribution}% <-this % stops a space
\thanks{$^{1}$ HB is with Carnegie Mellon University and FAIR, AI at Meta}%
\thanks{$^{2}$ AG, VK, and ST are with the Robotics Institute, Carnegie Mellon University, Pittsburgh, PA, USA}%
}

\begin{document}

\maketitle
\thispagestyle{empty}
\pagestyle{empty}

%%%%%%%%%%%%%%%%%%%%%%%%%%%%%%%%%%%%%%%%%%%%%%%%%%%%%%%%%%%%%%%%%%%%%%%%%%%%%%%%
\begin{abstract}
We pursue the goal of developing robots that can interact zero-shot with generic unseen objects via a diverse repertoire of manipulation skills and show how  passive human videos can serve as a rich source of data for learning such generalist robots. Unlike typical robot learning approaches which directly learn how a robot should act from interaction data, we adopt a factorized approach that can leverage large-scale human videos to learn how a human would accomplish a desired task (a human `plan'), followed by `translating’ this plan to the robot’s embodiment. Specifically, we learn a human `plan predictor’ that, given a current image of a scene and a goal image, predicts the future hand and object configurations. We combine this with a `translation’ module that learns a plan-conditioned robot manipulation policy, and allows following humans plans for generic manipulation tasks in a zero-shot manner with no deployment-time training. Importantly, while the plan predictor can leverage large-scale human videos for learning, the translation module only requires a small amount of in-domain data, and can generalize to tasks not seen during training. We show that our learned system can perform over 16 manipulation skills that generalize to 40 objects, encompassing 100  real-world tasks for table-top manipulation and diverse in-the-wild manipulation. \href{https://homangab.github.io/hopman/}{https://homangab.github.io/hopman/}
%\shubham{reword to emphasize generalization and zero-shot as opposed to only human videos}
\end{abstract}

\section{Introduction}
% \shubham{para 1 should setup the requirements. We need agents with diverse skills. moreover, these need to  generalize i.e. handle unseen objects. and, should do tasks zero-shot i.e. no online training for when deployed.}

A central goal in the rapidly growing area of robot learning is to develop generalist robots capable of performing a plethora of everyday manipulation tasks in diverse unseen real-world scenarios. In addition, to be practically useful, they should be able to accomplish these tasks out of the box when deployed in unseen scenarios. Towards this goal, our work pursues learning diverse core skills like manipulating articulated objects, picking,  placing, scooping, pouring, twisting, stacking, and swiping, among others that humans can effortlessly perform during everyday interactions. Moreover, we want these skills to be generalizable to unseen scenes with new objects, and be executeable in a ``zero-shot manner" i.e. without deployment-time training.

\begin{figure}[t]
    \centering    
    \vspace*{1em}
    \includegraphics[width=\columnwidth]{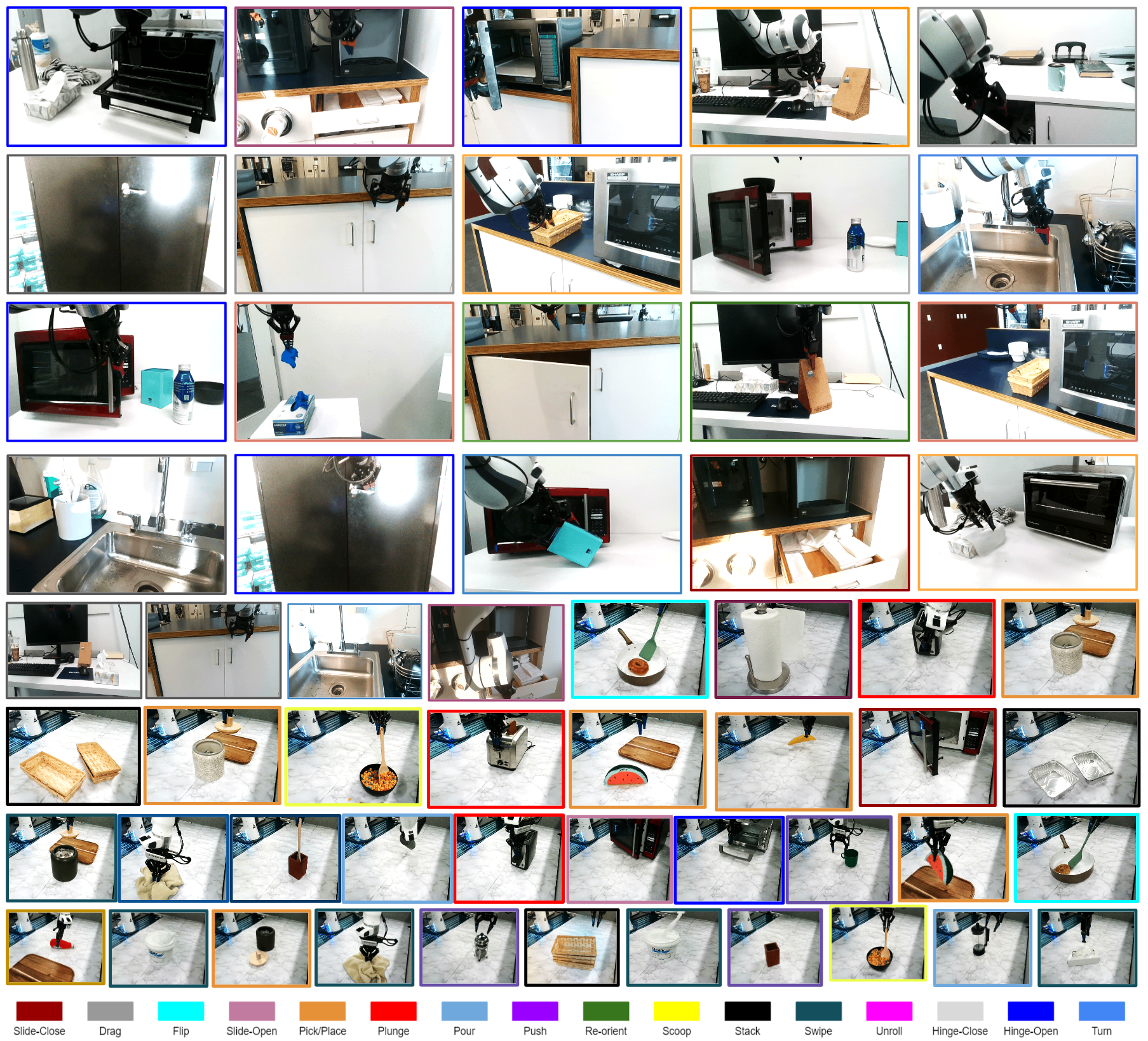}
    \caption{ A subset of different manipulation behaviors generated by our framework \algoName. By learning task-agnostic \textit{human-plan} prediction and \textit{robot-action} translation models, our system can interact with generic objects and execute diverse skills e.g.  unrolling, scooping, pouring, re-orientation, articulated object manipulation, etc. Videos are in the supplementary website \href{https://homangab.github.io/hopman/}{https://homangab.github.io/hopman/}}
    \label{fig:tasks} 
    \vspace*{-2em}
\end{figure}
An unsophisticated way to attempt this goal is to collect a gigantic robot interaction dataset for imitation learning. Albeit simple, this is not scalable for diverse real-world generalization because it would require collecting data not just for different tasks but for interaction across different objects with different skills, and is bottle-necked by physical access constraints. Indeed, recent approaches that attempt at developing diverse manipulation capabilities require years of on-robot data collection~\cite{rt1}, and are still largely limited to picking, placing, and pushing skills. Our solution is to factorize the task of learning a generalizable policy into 1) learning an interaction plan that captures changes that the object and the manipulator can undergo, 2) translate the plan into actions that can be executed on a robot. Our key insight is that the first module can leverage non-robot data, and in particular large passive datasets of human videos on the web. Given this \textit{human-interaction-plan}, acting in the real world reduces significantly in complexity as we only need to instantiate the human plan in a robot's context as \textit{robot-actions}. This translation model can be trained with limited paired human-robot data and generalizes to objects and scenarios that are unseen in the robot data since the human-interaction plan generalizes by virtue of diverse training.

\begin{figure*}[t]
    \centering    \includegraphics[width=0.85\textwidth]{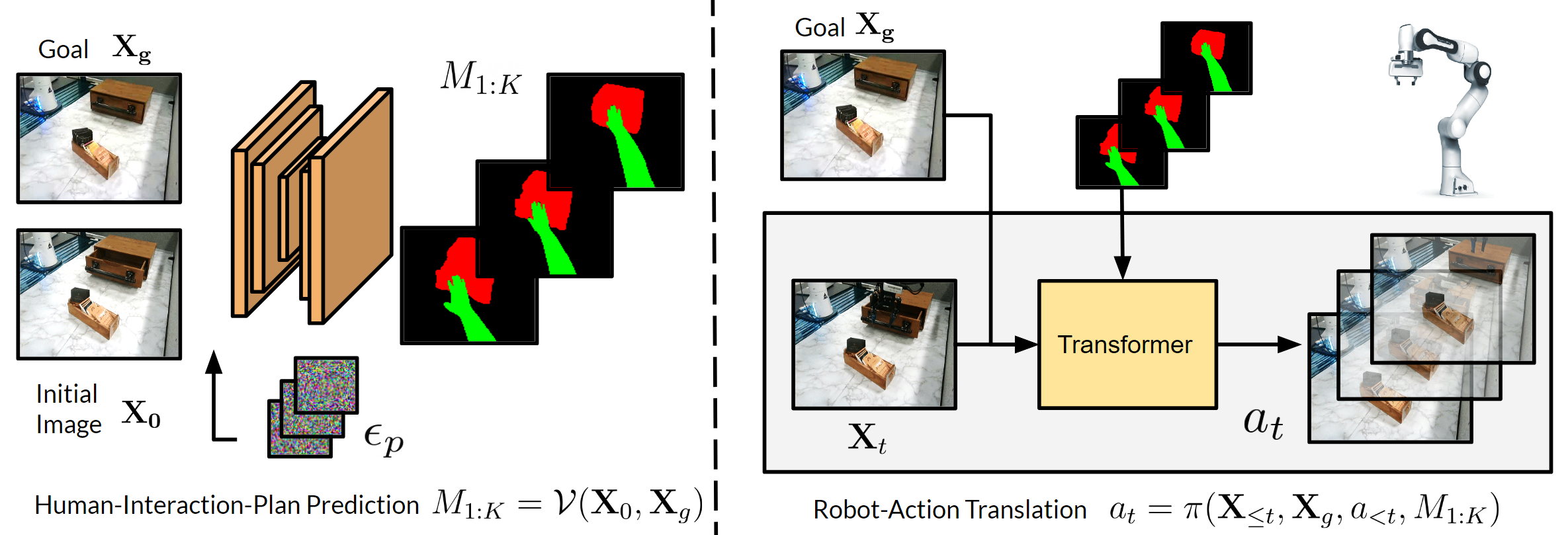}
    \caption{\footnotesize \algoName consists of a \textit{human-interaction-plan prediction model} (left), and a \textit{robot-action translation model} (right). Given an initial image of a scene $\mathbf{X_0}$ and a goal image $\mathbf{X_g}$, a diffusion model hallucinates plausible future hand and object masks $M_{1:K}$. These predictions along with current RGB observations of the scene $\mathbf{X}_t$ go as input to a translation model (instantiated as a closed-loop policy $\pi(\cdot)$) that outputs robot actions $a_t$ for executing the motions on a robot. Additional details on the approach are in section~\ref{sec:overall_framework}.}
    \vspace*{-1em}
    \label{fig:approach}
\end{figure*}

Some prior robot learning approaches have also investigated leveraging out-of-domain (human) data, primarily for learning visual representations~\cite{r3m,majumdar2023we,vip} and robotic affordances~\cite{nagarajan2019grounded,handsasprobes,bahl2022human, bahl2023affordances}.
However, these approaches require \textit{a lot of} further robot demonstrations for policy learning and typically also require a lot of deployment-time training. Other approaches learn task-specific action priors~\cite{qin2021dexmv,shawvideodex} for a few categories of manipulation tasks, with separate policies for each task. Compared to these, our approach of factorizing the overall policy can enable zero-shot manipulation over a range of diverse tasks, with a single policy that can be appropriately goal-conditioned and doesn't require any deployment-time training.

We consider semantic masks of hands and objects as a structured space for defining the \textit{human-plan}, since it abstracts out task-irrelevant details of the environment. Given an image of a scene and a goal image, we train the prediction model to predict the \textit{human-plan} as plausible future hand and object masks. We train this model across clips in diverse passive videos on the web and show that it generalizes to new scenes in our real-robot experiments. In order to transform the predictions to a physical embodiment's \textit{robot-actions} , we train a translation module on a small amount of paired data ($\sim$600 trajectories). We abbreviate our framework as \algoName~(\textbf{H}and \textbf{O}bject \textbf{P}lan for robotic \textbf{M}anipulation).

Through experiments on a set of 100 tasks, involving 16 skills and 40 objects, we show \algoName~ can help distill information about manipulation from passive human videos on the web to physical scenes in a robot's workspace, as evaluated through generalization across five different axes. In summary, we make the following contributions:

\begin{itemize}[leftmargin=*]
    \item Present an approach for learning goal-conditioned prediction of hand-object interaction plans using everyday interaction videos.
    \item Develop a framework that casts robot manipulation as translation of (predicted) hand-object plans, thus allowing the use of easily available human videos for learning diverse manipulation. 
    \item Demonstrate the overall framework across 100 manipulation tasks involving 40 objects with 16 skills, while evaluating generalization in a structured manner for table-top manipulation and in-the-wild manipulation in unseen scenes. 
\end{itemize}

\section{Related Works}
\vspace{-0.2em}

\noindent\textbf{Understanding human interactions from videos.} Several recent approaches in computer vision have focused on understanding hand-object interactions in diverse everyday settings~\cite{smthsmth,youcook,cmudata,ego4d,Shan20,epic,egtea}. Specifically, prior work has investigated learning hand pose estimation~\cite{zimmermann2017learning,iqbal2018hand,spurr2018cvpr,ge20193d,baek2019pushing,boukhayma20193d,hasson2019learning,dkulon2020cvpr,liu2021semi,rong2020frankmocap}, %%% trim this to hand-object and hand papers
object pose estimation~\cite{Kehl2017SSD6DMR,Rad2017BB8AS,Xiang2018PoseCNNAC,Hu2019SegmentationDriven6O,He2020PVN3DAD}, interaction hotspot prediction~\cite{nagarajan2019grounded,liu2022joint,handsasprobes}, prediction of plausible hand grasps~\cite{mo2021where2act,brahmbhatt2019contactgrasp}, and activity understanding~\cite{activity1,activity2}.  Our human-interaction-plan prediction module is inspired by these developments, where we focus on learning motions of hands and objects from passive human videos that are directly relevant for manipulation, and abstract out task-irrelevant visual details through semantic masks.

\noindent\textbf{Learning Visual Representations for Manipulation.} A growing body of recent works learn mappings from visual observations to robot actions for performing tasks~\cite{visual_imitation1,visual_imitation2,visual_imitation3}. One common way of using data beyond robot interactions for efficient learning is to pre-train the visual representations which serve as backbones for the policy models~\cite{r3m,vip,majumdar2023we,pvr,pvr2} with passive human videos~\cite{ego4d,kay2017kinetics} and image data~\cite{imagenet}. However, these methods still crucially rely on a lot of in-domain robot data or deployment-time training, and are  restricted to learning task-specific policies.

\noindent\textbf{Learning Affordances.} Towards learning structure more directly related to manipulation, some works try to predict visual affordances in the form of where to interact in an image, and local information of how to interact~\cite{mo2021where2act,handsasprobes,bahl2023affordances,liu2022joint}. While these could serve as good initializations for a robotic policy, they are not sufficient on their own for accomplishing tasks, and so are typically used in conjunction with online learning, requiring several hours of deployment-time training and robot data~\cite{bahl2022human,bahl2023affordances}. Our work differs from this in terms of predicting an approximate motion of how a human hand and the object is likely to move for the entire trajectory (not just at/near contacts unlike affordances) and is \textit{zero-shot} in terms of not requiring any deployment-time training.

\noindent\textbf{Manipulation without deployment-time training.} 
With a goal similar to ours of using human videos to learn models that can be directly deployed, some approaches leverage curated data of human videos~\cite{qin2021dexmv,shawvideodex} for learning task-specific policies (instead of a single model across generic tasks). Others that train a single policy across tasks require large in-domain perfectly aligned human-robot data~\cite{wang2023mimicplay,smith2019avid,xiong2021learning} and are not capable of leveraging passive web data. Perhaps most closely aligned with ours, Bharadhwaj \etal ~\cite{bharadhwaj2023zero} learn (human) action trajectories from passive web videos and leverage a heuristic to convert these to robot trajectories. However, their actions are restricted to simple coarse open-loop motions that do not involve  grasping, and hence can't exhibit diverse skills.  Compared to these, our framework utilizes \textit{diverse large-scale} passive human video data on the web, combined with a \textit{small amount} of in-domain robot data, with a single model capable of tackling different manipulation tasks zero-shot.

\section{\textbf{H}and-\textbf{O}bject \textbf{P}lan for robotic \textbf{M}anipulation}
\label{sec:overall_framework}

We aim to develop a robot manipulation system that can accomplish diverse skills zero-shot with a plethora of different unseen objects in the real world. Our key insight is to leverage a factorized policy model (see Fig.~\ref{fig:approach}) that consists of two stages: a) a goal-conditioned human plan prediction model that predicts future masks for plausible hand and object motions, and b) a translation model that learns to transform the corresponding predicted plans into actions that can be executed with a robot for real-world manipulation. We show how we can train the human-plan prediction model on diverse passive human videos from existing large scale datasets, and use it for predicting plausible plans in a robot's environment. In contrast, the translation model can be trained with a small amount of paired human-robot data. This factorization allows us to generalize to scenarios that are unseen in the robot data, because the human-interaction-plan model with its diverse training generalizes well, and the translation model is tasked with a simpler job of converting these plans to the robot's embodiment.

\subsection{The Human-Plan Prediction Model}
\label{sec:fm}
Instead of predicting the future in the image space, we focus on predicting only the motion of the human hand and the object being interacted with, in terms of respective semantic masks. We enable this prediction through a diffusion model trained on diverse human videos on the web. For each video in the training data, we extract hand-object masks for each frame. Let $\mathbf{M_{1:K}}$ denote the respective mask frames from time steps $1$ to $K$. For simplicity we consider each mask frame to be an image, where all the hand pixels are green, all the object pixels are red, and the rest of the pixels are black. Let $\mathbf{X_0}$ denote the first frame (RGB) of the video, $\mathbf{X_g}$ denote the last frame (RGB) of the video, which will act as a goal frame, and $\mathcal{V}(\mathbf{X_0,X_g})$ denote the prediction model. 
%For simplicity, let $\mathbf{X_c}$ denote the conditioning frames, such that $\mathbf{X_c}=\{\mathbf{X_0,X_g}\}$. 
\begin{figure}[h!]
    \centering    \includegraphics[width=\columnwidth]{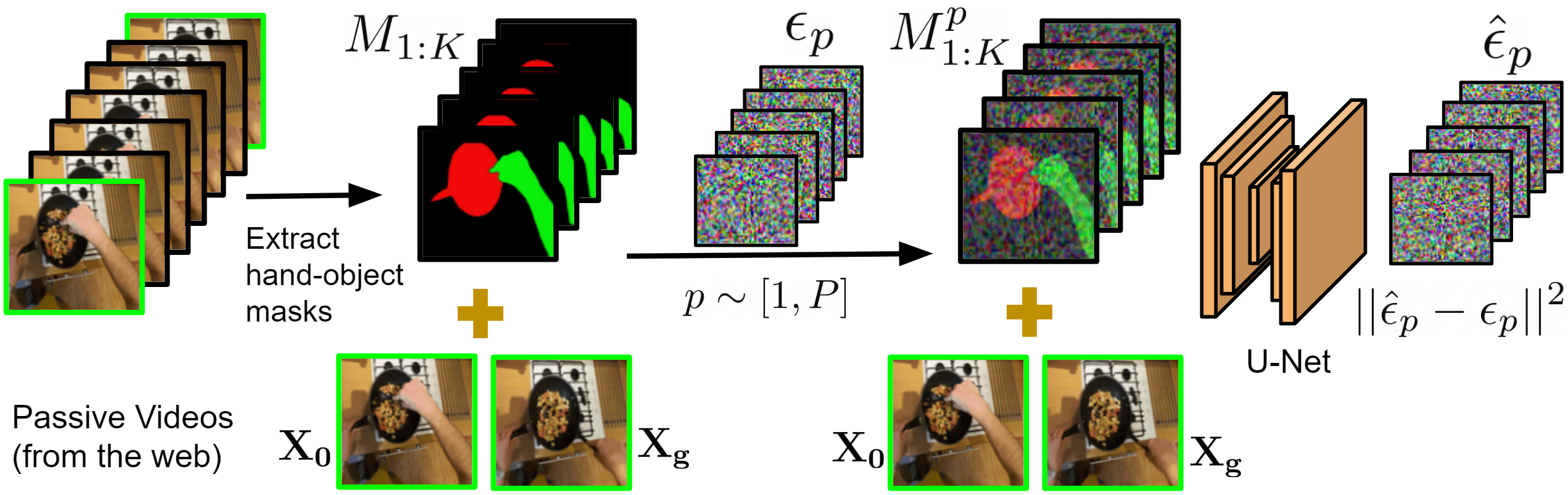}
    \caption{\footnotesize Detailed illustration of a training pass through the future prediction model. This is a diffusion model, with a U-net that predicts per-frame noise at each step $p$ of the diffusion process. Additional details on the model and training are in Section~\ref{sec:fm}.}
        \vspace*{-1em}
    \label{fig:diffusion}
\end{figure}
In the forward diffusion process, all the mask frames $\mathbf{M_{1:K}}$ are corrupted by incrementally adding noise, and converging to a unit Gaussian distribution $N(\mathbf{0},\mathbf{I})$. New samples can be generated by reversing the forward diffusion process, by going from Gaussian noise back to the space of mask frames. To solve the reverse diffusion process, we need to train a noise predictor $\epsilon_\theta(\cdot |t)$ which is a time-conditioned U-net~\cite{stablediffusion,voleti2022masked} trained to predict the noise at each step of the diffusion process. The input to the network at step $t$ of the diffusion process is a channel-wise concatenation of the conditioning frames and noisy mask frames $\mathbf{[\mathbf{X_0,X_g},M_{1:K}^t]}$, the output is the predicted noise of same dimensionality as the input. Fig.~\ref{fig:diffusion} illustrates this visually, and equation 1 shows the training objective $\mathcal{L}(\theta)$.
\begin{align*}
&\mathbb{E}_{t,\mathbf{[\mathbf{X_0,X_g},M_{1:K}]}\sim p_{\text{train}},\mathbf{\epsilon}\sim\mathcal{N(\mathbf{0},\mathbf{I})}}\\
&\left[||\epsilon - \epsilon_\theta(\sqrt{\overline{\alpha}_t}\mathbf{M_{1:K}} + \sqrt{1-\overline{\alpha}_t}\mathbf{\epsilon} | \mathbf{X_0,X_g},t)||^2 \right]
\end{align*}

Here $\overline{\alpha}_t$ is a hyper-parameter that depends on the noise schedule of the diffusion process. During inference, given $\mathbf{X_0,X_g}$ we obtain $M_{1:K}=\mathcal{V}(\mathbf{X_0,X_g})$ through reverse diffusion.

\subsection{The Robot-Action Translation Model}
\label{sec:translation}
\begin{figure}[h!]
    \centering
    \vspace{-0.5em}
\includegraphics[width=0.95\columnwidth]{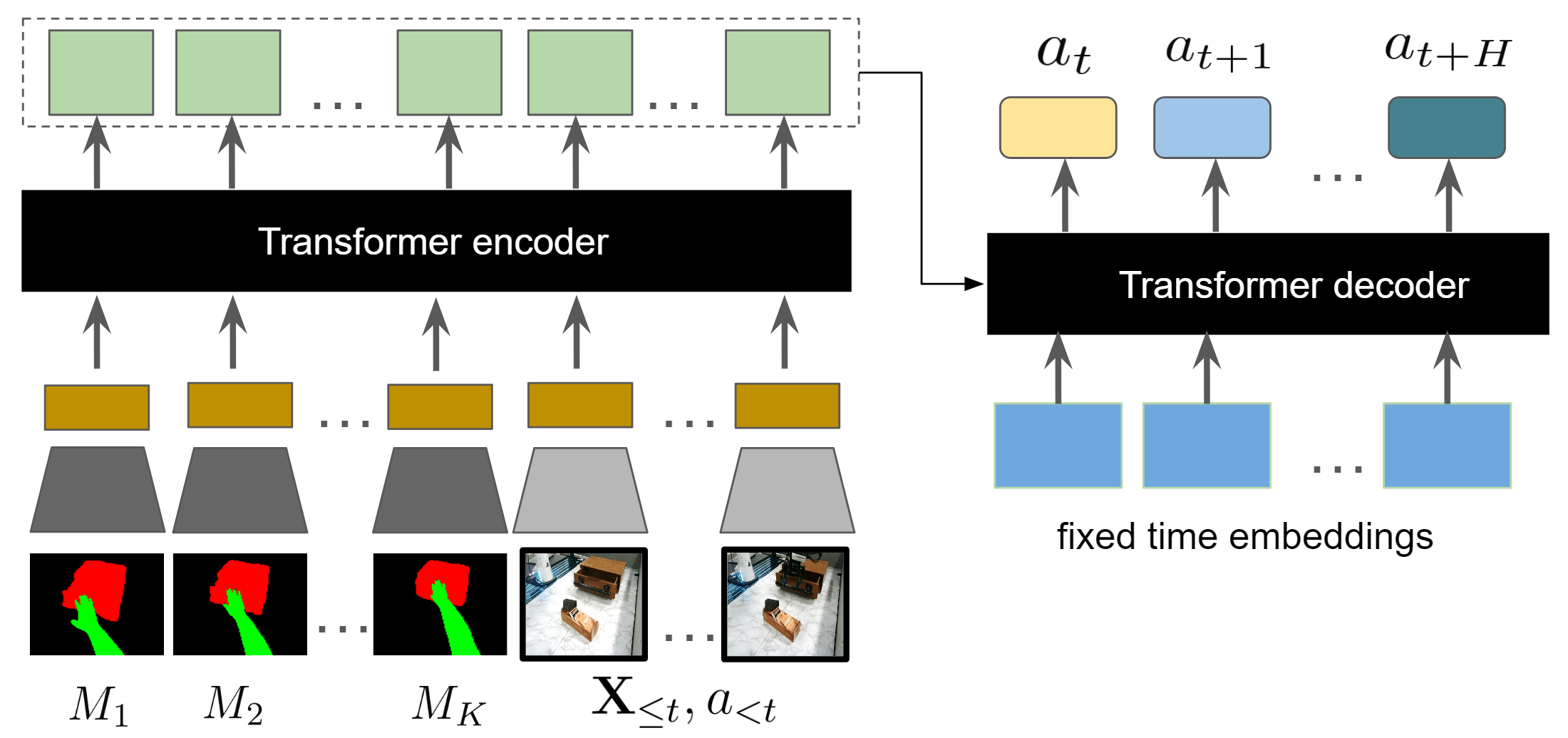}
    \vspace{-0.7em}
    \caption{\footnotesize Architecture of the translation model that transforms predicted future hand-object masks to a robot trajectory, described in section~\ref{sec:translation}}
        \vspace{-1em}
    \label{fig:translation}
\end{figure} 

We use the human-plan predictor discussed in Section~\ref{sec:fm} to hallucinate plausible future hand and object masks for interaction in a robot's physical scene. However, this human-plan doesn't directly inform what actions the robot should execute to be able to perform the desired interaction. To enable robot manipulation in the context of the predicted plans, we learn a translation model. The translation model is a transformer that is conditioned on the outputs of the future prediction model $M_{1:K}$ and for each observation $\mathbf{X}_t$, and predicts actions for $H$ steps in the future. The model behaves as a closed-loop policy $\pi(\mathbf{X}_{\leq t},\mathbf{X}_{g},a_{< t},M_{1:K})$ that is queried at each time-step $t$ during deployment. Predicting multiple time-steps $H$ in the future and averaging actions during deployment, helps in executing smooth robot motions, with less compounding errors~\cite{act}. We describe the architecture of the translation model in Fig.~\ref{fig:translation} and additional details in Appendix A.3.

For training the translation model, we need some paired human-robot data, where we have pairs of trajectories that involve a robot manipulating an object, and a human manipulating a similar object. To obtain such paired trajectories, we develop two approaches:

  \begin{figure}[h!]
    \centering
\includegraphics[width=\columnwidth]{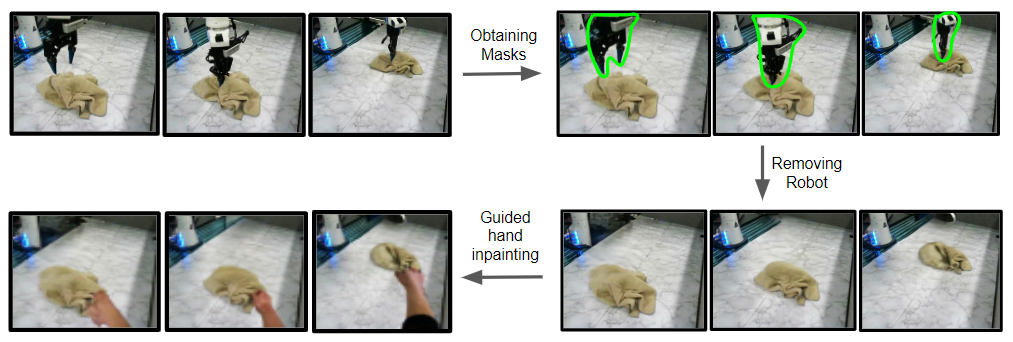}
    \vspace{-1.5em}
    \caption{\footnotesize Illustration of the different steps in generating hallucinated human hand trajectories from robot trajectories. This is an alternate data source for the translation model in addition to collecting paired human-robot data.}
    \vspace{-1em}
    \label{fig:inpainting}
\end{figure} 
 \textbf{Collecting paired demonstrations:} A human operator tele-operates a robot in scene, and after reset, or in a parallel identical setup, a human manipulates a similar object with an approximately similar motion as the robot arm. Collecting this paired data is not very expensive, and we spent around 3 days to collect 600 trajectories.

 \textbf{Hallucinating paired data:} 
To augment the paired demonstrations, we also propose to leverage (more easily collectable) robot-only data. To  obtain hallucinated pairs, we can  convert videos of a robot trajectory into a videos of a human trajectory  through recent advances in hand in-painting techniques~\cite{ye2023affordance,stablediffusion} . Specifically, we obtain robot masks per frame through simulation, and perform inpainting to remove the robot from the scene. We then perform guided in-painting of a plausible human hand~\cite{ye2023affordance} around the location of the robot end-effector in the scene. Fig.~\ref{fig:inpainting} visually illustrates this process of hallucinated data generation. In the experiments, we show how hallucinated paired data generated through this approach can be used to boost the performance of the translation model.

\vspace{-0.2em}
\section{Experiments}
\vspace{-0.2em}
Through experiments with diverse real-world objects in unseen scenarios, we demonstrate generalization of our framework for several robot manipulation tasks. Videos are in the website \href{https://homangab.github.io/hopman/}{https://homangab.github.io/hopman/}
\vspace{-0.3em}

\subsection{Experiment Settings}
We consider two different types of manipulation settings for experiments - table top scenarios with a fixed robot and camera, and in-the-wild manipulation with the same robot and camera on a mobile base. 

\noindent\textbf{Table-Top Manipulation.} We consider several everyday objects with different plausible manipulations for our experiments. We demonstrate results on a total of 16 skills: pouring, plunging, pushing, picking/placing, slide-opening, slide-closing, hinge-opening, hinge-closing, swiping, dragging, flipping, scooping, in-place re-orientation, unrolling, and stacking, and  40 object types, with 2-3 instantiations per object type, comprising around 100 tasks.  Detailed list of objects and tasks are in the Appendix section A.2

\noindent\textbf{In-the-Wild Manipulation.} We drag a Franka Panda arm on a mobile base across natural kitchen and office scenes. The camera is also attached to the base, and moves along with it. For these experiments we fine-tune the translation model used for the table-top experiments, on $\sim$200 additional paired trajectories collected with the mobile robot. For evaluation, we consider the same generalization levels described above. This setting is much more challenging because in addition to object and skill variations, we also have scene variations, including completely new scenes never seen in the paired data. Details of variations are in the supplementary website.

\subsection{Training data}
\label{sec:overall_setup}
 The training data for our framework consists of a large set of passive web videos, a small amount of paired human-robot in-domain data, and some unpaired robot-only data.

\noindent\textbf{Passive Human Data.} For the future prediction model, we use existing passive human videos~\cite{visor,ego4d,smthsmth} and obtain ground-truth semantic masks for the right hand and the object being interacted with the right hand in each frame~\cite{visor,hos}. We sample short video clips, each lasting a few seconds and do not curate the videos in any way with tasks or language labels. Details about ground-truth mask extraction for different datasets are mentioned in the Appendix A.3

\noindent\textbf{Paired Data.} For the translation model, we use a small amount of paired collected by us ($\sim$400 trajectories in-lab and $\sim$200 trajectories in-the-wild) and a larger robot-only data ($\sim$1000 trajectories) combined with hallucinated hand masks through the approach described in section~\ref{sec:translation}. All the robot data are collected through an adaptation of the tele-operation stack proposed in~\cite{kumar2015mujoco}.

\subsection{Defining Tasks and Evaluating Generalization}
\label{sec:definitions}

\begin{figure}
    \centering
    %\vspace{-1em}
\includegraphics[width=\columnwidth]{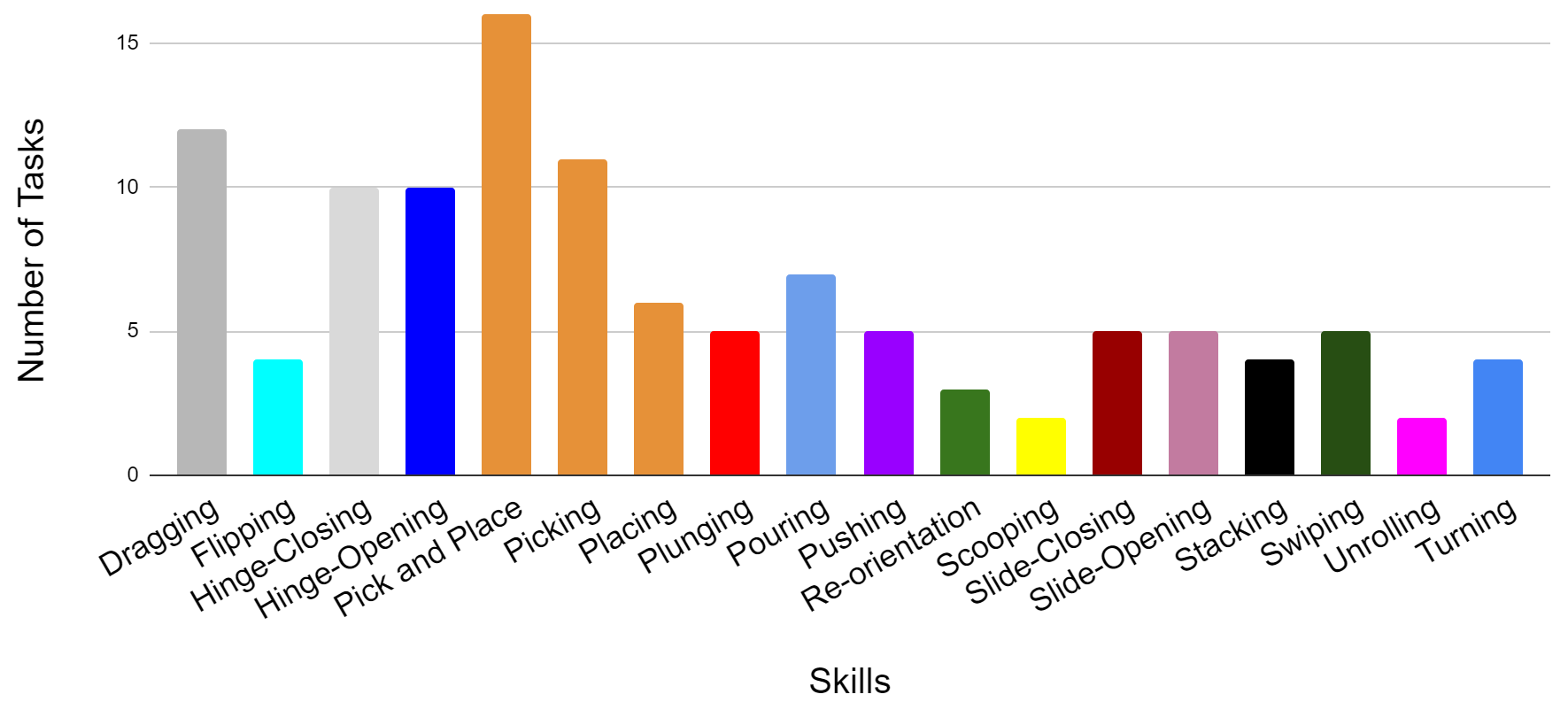}
    \vspace{-1.5em}
    \caption{\footnotesize Distribution of skills across tasks in our experiments. The diversity of skills is more representative of real-world distributions, compared to pushing/pick and place that is predominant in robot learning papers.}
    \label{fig:skilldistribution}
    \vspace{-1.5em}
\end{figure} 
\begin{figure*}[t]
    \centering
\includegraphics[width=0.95\textwidth]{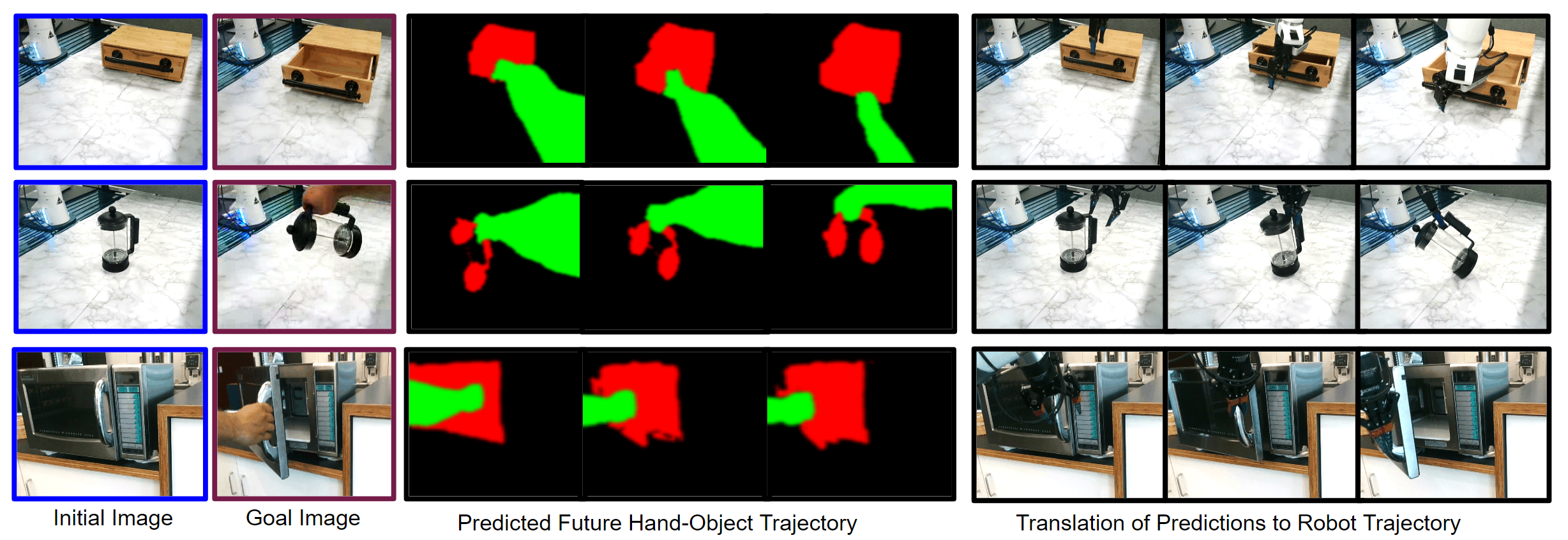}
    \caption{\footnotesize \textbf{Qualitative results for the entire framework.} We show qualitative results for the predicted hand-object trajectory given an initial image of a scene and a goal image, followed by translation of the predictions to a robot trajectory for execution in the real world.}
    \label{fig:qualvis}
\end{figure*}
\begin{figure*}[t]
    \centering
\includegraphics[width=0.95\textwidth]{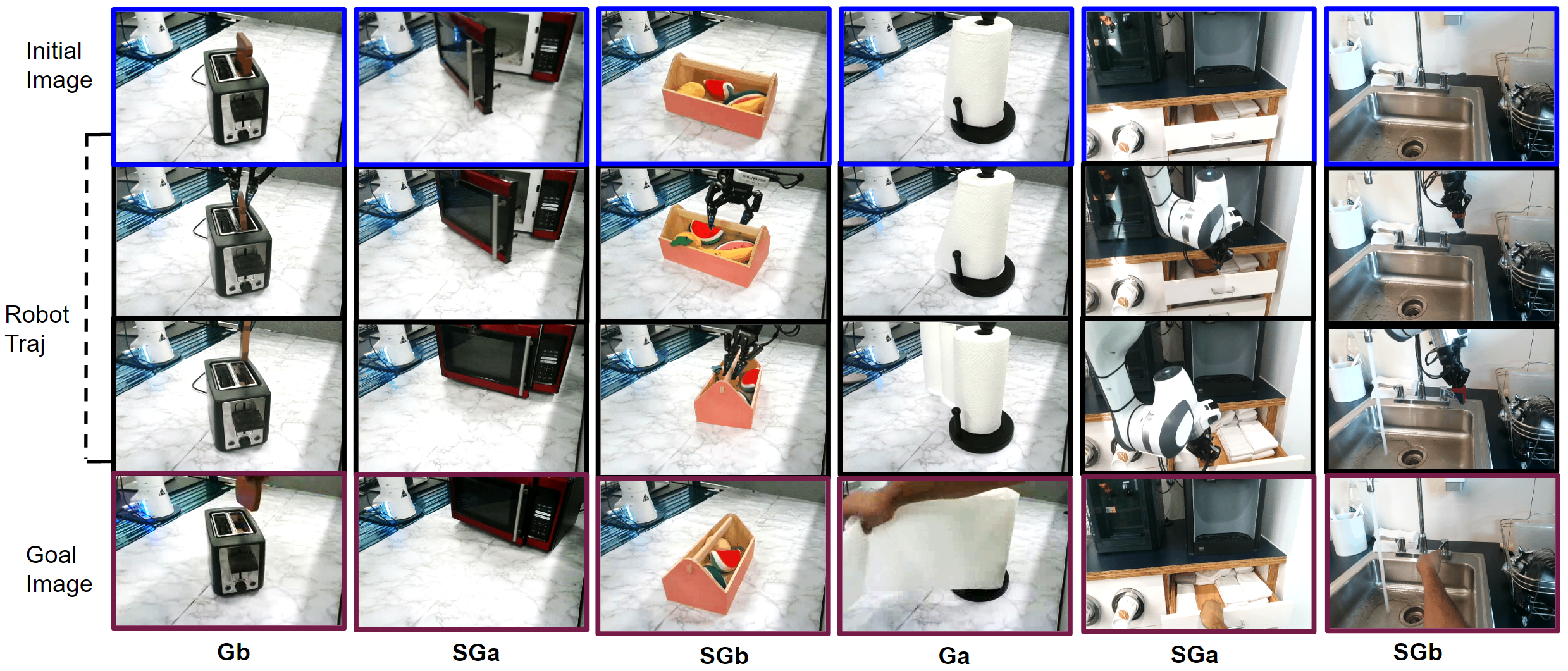}
    %\vspace{-1.5em}
    \caption{\footnotesize \textbf{Examples of robot evaluations.} We show qualitative results for robot evaluations, with an intermediate image and the image corresponding to the final state reached by the robot, for a given initial scene and a goal image. Subscripts show the type of generalization for each evaluation, as described in sec~\ref{sec:definitions}. More robot videos of evaluations are in the linked website.}
    \label{fig:qualvisrobot}
        \vspace{-1.5em}
\end{figure*}

Prior works in robot learning adopt widely different and oftentimes inconsistent definitions of generalization criteria. Some prior works~\cite{moo,rt1,cacti,qin2021dexmv} consider seen vs. unseen objects, where the unseen objects often involve different instantiations of the seen objects, with shape, color, and texture variations, with skills (e.g. pushing, picking etc.) that are always seen in the training data. Others~\cite{haldar2023watch,cui2022play} only consider generalization in terms of position and configuration variations of seen objects.  In light of this, in this paper, we develop a structured criteria for evaluating generalization in terms of object categories, object instantiations, object configurations, and skills. We adopt the following definitions

\begin{itemize}[leftmargin=*]
\item \textbf{Task definition:} Each task is a tuple consisting of (\objectcategory, \instantiation, \skill). Here, \textit{\objectcategory} denotes the type of the object e.g. `drawer', `mug', `toaster' etc. While, \textit{\instantiation} defines a particular object within a category, with a specific instantiation of color, shape, size, and texture. Finally, \textit{\skill} defines the particular behavior e.g. `open', `flip', `push' etc. that can be done with an object. 
\item \textbf{Mild generalization (\wG):}  This involves generalizing among unseen configurations (i.e. position and orientation variations) for seen \instantiation~ and seen skills, along with mild variations in the scene like lighting changes.
     \item \textbf{Standard generalization (\sG):} We have the following types of generalization in this category
     \begin{itemize} 
      \item \textbf{instance generalization (\sG \texttt{a}):} In addition to variations in \wG, in \sG \texttt{a} we evaluate unseen \instantiation~ for seen skills. For example, only a red mug is seen with the push skill in training, and we generalize to pushing motions for green, and purple mugs of different shapes and textures. 
    \item \textbf{unseen combinations (\sG \texttt{b}):}  This includes scenarios with unseen (\objectcategory, \skill) pairs but each seen independently in training. So atleast one instance of an \objectcategory~ is seen during training, and the \skill is also seen during training but not in relation to this object. For example, `open' is seen, and `close door' is seen but `open door' is not seen in training. 
\end{itemize}
     \item \textbf{Strong Generalization (\sTG):}  We categorize the following types of generalization that involve either a completely unseen \objectcategory~ or an unseen \skill~ into this category. These are very challenging tests of generalization.
     \begin{itemize} 
        \item \textbf{\objectcategory~ completely unseen (\sTG \texttt{a}): } This includes scenarios where a particular \objectcategory~ e.g. microwave is never seen in training
        \item \textbf{\skill~ completely unseen (\sTG \texttt{b}):} This includes scenarios where a particular \skill~ e.g. re-orientation is never seen in any context during training. 
     \end{itemize}
\end{itemize}

Note that our formalization of generalization is centered around objects being interacted with and the skills that are possible for interaction, and we do not consider scene variations of the background in the definitions, unlike some prior work~\cite{genaug,rosie,rt1,cacti,bharadhwaj2023roboagent}. However, for experiments, we consider diverse scenes, both for table-top manipulation and manipulation of objects in-the-wild in unseen kitchens and offices.
\vspace{-0.2em}
\subsection{Baselines and Ablations:}\vspace{-0.2em}
We consider a goal-conditioned behavior cloning baseline (BC) trained on all the robot data ($\sim$1600 trajectories). The architecture of the policy is a transformer similar to our translation model without the conditioning on human-interaction-plans. The next baseline (MP) uses paired human-robot data, and is an adaptation of~\cite{wang2023mimicplay}. We compare with VRB~\cite{bahl2023affordances} by using the affordance model from the paper to do affordance conditioned imitation learning. We also consider a baseline that is trained entirely with passive human videos, for coarse manipulation (H2R)~\cite{bharadhwaj2023zero}. In addition to these, we consider variations of our translation model trained on only in-lab paired human-robot data ($\sim$400 trajectories), only hallucinated data ($\sim$1000 trajectories), and combined paired and hallucinated data ($\sim$1400 trajectories).

\vspace{-0.2em}
\subsection{Evaluating Goal-conditioned Manipulation}
\vspace{-0.2em}
\begin{figure}[t]
    \centering
\includegraphics[width=\columnwidth]{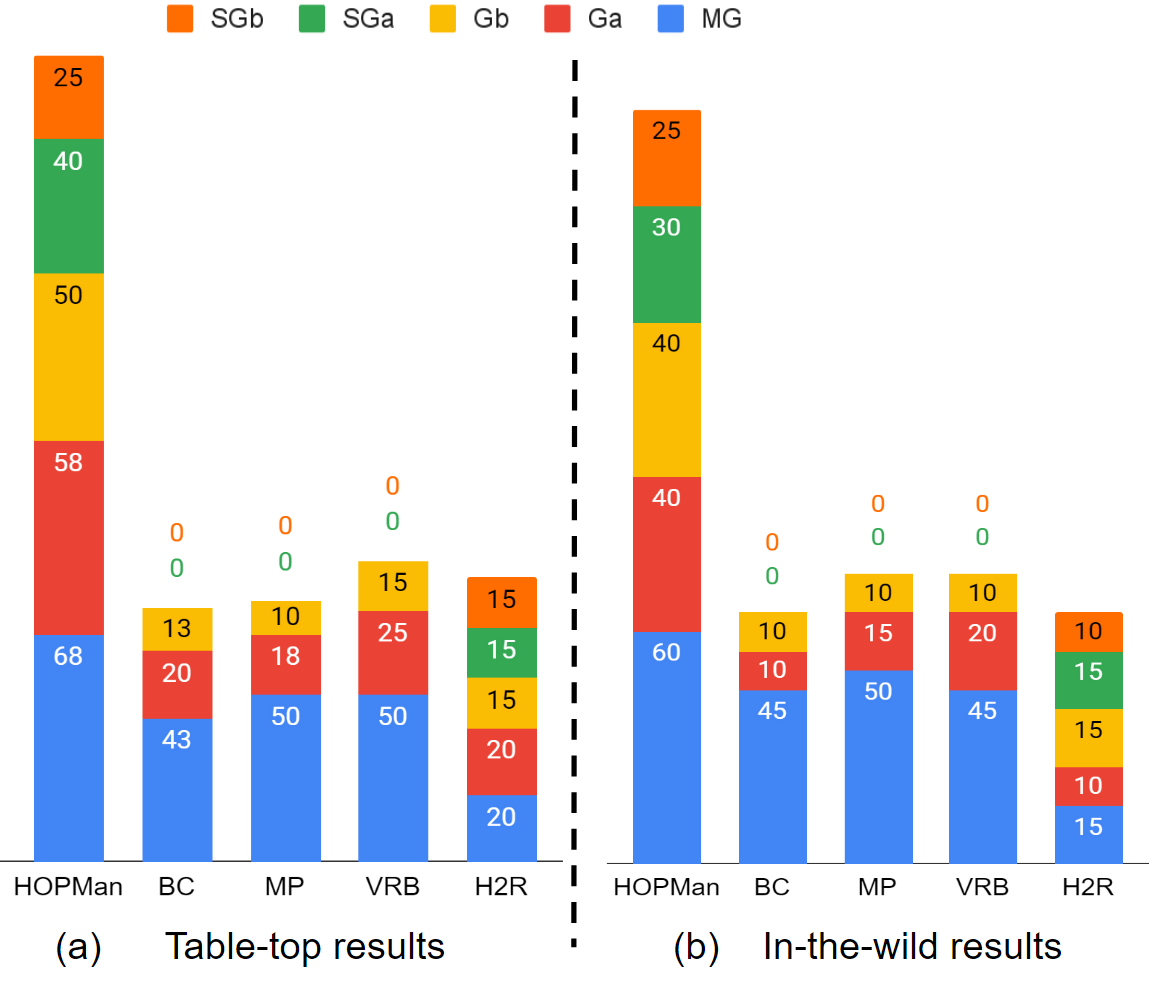}
    \vspace{-2em}
    \caption{\footnotesize \textbf{Summary of results.} The numbers represent success rates for goal-conditioned evaluations, in terms of \% of trials that correspond to manipulating objects in the scene to bring them to the desired goal configurations. We perform evaluations separately for the table-top manipulation and in-the-wild manipulation experiments.}
    \label{fig:genresults}
    \vspace{-1.5em}
\end{figure}

\setlength{\tabcolsep}{4pt}
In this section, we evaluate \algoName for robot manipulation. Given an image of a scene in the robot workspace and a goal image, we use the human-interaction-plan predictor to output a sequence of plausible hand-object masks, which are input to the translation model that performs closed-loop control for executing a sequence of actions on the robot. We evaluate across diverse unseen objects exhibiting several plausible skills, and unseen scenes in-the-wild, and tabulate success rates by aggregating over objects for each skill. We define success in terms of whether the object is brought to the desired configuration in the goal image.

Fig.~\ref{fig:qualvis} shows qualitative results for \algoName where we see that the generated human-interaction-plans are plausible and correspond to manipulating the object to obtain the specified goal configuration. In Fig.~\ref{fig:qualvisrobot} we show more robot evaluations in terms of an intermediate frame in the trajectory and the final frame reached at the end of robot evaluation, for different initial and goal images. 

In Fig.~\ref{fig:genresults} we summarize quantitative evaluations across the different generalization axes. For standard generalization \sG~ and strong generalization \sTG, we see that \algoName achieves significantly high success rate. This demonstrates the effectiveness of learning plausible manipulation trajectories of hands and objects from internet videos combined with small paired data, for generalization to diverse settings, in comparison to relying on only in-domain data (BC, MP baselines), on predicting visual affordances combined with robot data (VRB) or on only passive data (H2R). 

\vspace{-0.2em}
\subsection{Ablations of the Translation Model}
\begin{figure}[t]
    \centering
\includegraphics[width=\columnwidth]{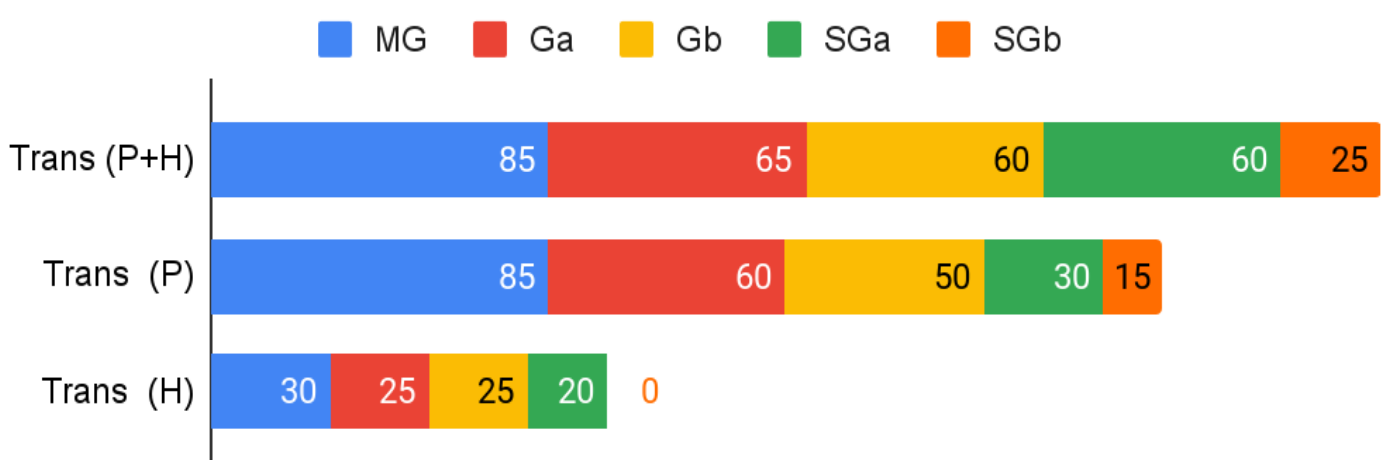}
     \vspace{-2em}
    \caption{\footnotesize \textbf{ Translation model ablations.} Ablation results for the translation model alone with specified masked hand-object trajectories instead of future predictions. Here, P denotes paired data, and H denotes hallucinated data, described in section~\ref{sec:translation}. and the numbers represent success rates.}
    \label{fig:transresults}
    \vspace{-1.5em}
\end{figure}

\vspace{-0.2em}
In this section, we evaluate the translation model in isolation independent from the prediction model. Specifically we evaluate how good is the translation model in translating the motion of a ground-truth hand-object trajectory into robot trajectories. Here, we introduce different objects in the scene and manually execute a motion with a human hand to reach the goal, and then pass the video through the hand-object segmentation model. We ablate over three variations of the the translation model, trained with paired data and hallucinated data, trained with only paired data, and trained with only hallucinated data, in table-top settings. From Fig.~\ref{fig:transresults}, we observe that training the model with combined paired and hallucinated data (P+H) leads to better performance than training with just paired data (P) indicating that the translation model is able to effectively utilize imperfect hallucinated trajectories for improving generalization.

\vspace{-0.2em}

\section{Discussion and Limitations}
\vspace{-0.2em}
In this work, we developed a framework for learning generalizable robot manipulation by combining internet-scale human videos of everyday interactions with limited in-domain robot demonstrations. Leveraging these, our framework can accomplish diverse tasks by predicting plausible hand-object plans and translating these to the robot's embodiment.
%We show how it is possible to leverage internet-scale video data of human interactions to train the future prediction model, and use only very limited in-domain paired human-robot data for training a translation model.
% We hope our work demonstrates a promising direction for learning generalizable robot manipulation skills by extracting useful abstractions from passive human videos, beyond visual representations. 
Broadly, our work is indicative of how rich out-of-domain datasets like human videos can alleviate the data paucity that greatly bottlenecks robot learning by helping learn hand-object interaction plans, and enable wide generalization of manipulation skills to unseen scenarios. While our framework does allow strong generalization to unseen tasks, these are still limited in their complexity and it would be an interesting future direction to extend our approach for tacking long-horizon tasks that requiring composing multiple skills. Moreover, our framework may struggle with dexterous manipulation tasks as recovering precise hand and finger articulations from web videos remains a challenge in computer vision. %Lastly, our framework requires (albeit limited) in-domain data for learning the translation model, and further relaxing this requirement would be an interesting direction.
%Future work could address limitations like requiring (albeit very small) paired human-robot data, and short prediction horizon in order to develop versatile agents that learn long-horizon manipulation behaviors comprising of several chained skills.

\clearpage
\newpage
\footnotesize
\bibliographystyle{IEEEtran}
\bibliography{example}  % .bib

\newpage
\clearpage
\onecolumn
 \normalsize
\appendix
\section{Supplementary}

\subsection{Robot Evaluation Videos}
Robot videos are in the website \href{https://homangab.github.io/hopman/}{https://homangab.github.io/hopman/}
\subsection{List of Tasks}
\label{sec:tasks}
% \noindent\textbf{Example skills:} pushing, picking, pick and place, plunging, pouring, opening, closing, unrolling, swiping, dragging, flipping, grasping, reorienting in place....

% \noindent\textbf{Example objects:} toaster oven (horizontal hinge), drawer (planar motion), microwave oven (vertical hinge), tea bags, tray, strainer, toaster, ketchup bottle, watermelon/fruit slices, cloth, tissues in a tissue box, paper towel roll, cup, mug, bowl, french press, Spice box

% \noindent\textbf{Example object instance variations:} Shape, texture, color,  and combinations of these. Some objects have variations large enough to be classified as \textit{different} objects (e.g. bowls, fruit slices, some drawers and ovens), Tool Box 

% Tuples below in the format (Object, Instance, Motion):

% SEEN (Have paired data for these)
% \begin{itemize}
%     \item (Toaster, Bowl1, Plunging)
%     \item (Toaster, 1, Picking Toast)
%     \item (Drawer, 1, Opening)
%     \item (Drawer, 1, Closing)
%     \item (Toaster Oven, 1, Opening)
%     \item (Toaster Oven, 1, Closing)
%     \item (French Press, 1, Pouring)
%     \item (Cloth, 1, Swiping)
%     \item (Tissue Box, 1, Picking)
%     \item (Tool Box, 1, Pick and Place)
%     \item (Ketchup Bottle, 1, Pick and Place)
%     \item (Spice Box, 1, Pick and Place)
%     \item (Bowl, 1, Pushing)
%     \item (Bowl, 2, Pushing)
%     \item (Bagel, 1, Flipping)
%         \item (Tissue Roll, 1, Unrolling)
%             \item (Mug, 1, Flipping)    \item (Bagel, 1, Flipping)
% \end{itemize}
\begin{figure}[h!]
    \centering    \includegraphics[width=0.4\textwidth]{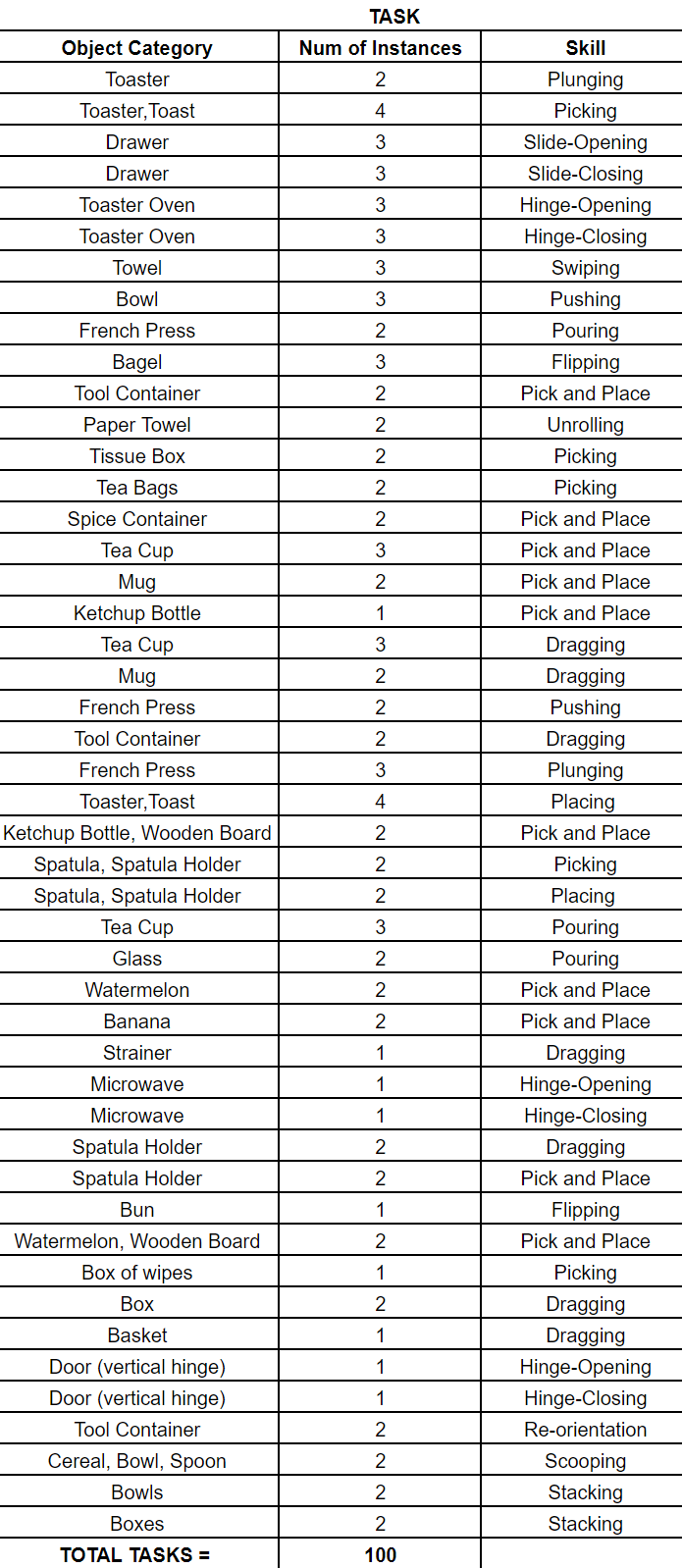}
    \caption{Summary of the different tasks for table-top manipulation experiments in terms of object types, number of instantiations per object type (variations in shape, size, color ,texture) and verbs denoting the type of possible skill with each object type}
    \label{fig:tasks}
\end{figure}

\subsection{Additional details on the models}

\subsubsection{Human-Plan Prediction model:}
Instead of predicting the future in the image space, we focus on predicting only the motion of the human hand and the object being interacted with, in terms of respective semantic masks. We enable this prediction through a diffusion model trained on diverse human videos on the web. For each video $\mathcal{V}$ in the training data, we extract hand-object masks for each frame . Let $M_{1:K}$ denote the respective mask frames from time steps $1$ to $K$. We set the value of $K=7$ for our experiments, which amounts to choosing 7 uniformly space frames in a 2 second window of a video clip.  For simplicity we consider each mask frame to be an image, where all the hand pixels are green, all the object pixels are red, and the rest of the pixels are black. the Let $X_0$ denote the first frame (RGB) of the video, and $X_g$ denote the last frame (RGB) of the video, which will act as an optional goal frame. The diffusion model operates at a resolution of 64x64 for the predicted masked frames.

We train two version of the future prediction model: 1) \textit{unconditional prediction} that is conditioned on only $\mathbf X_0$, and 2) \textit{goal-conditioned prediction} that conditioned on both $\mathbf X_0$ and $\mathbf X_g$. In the forward diffusion process, all the mask frames $M_{1:K}$ are corrupted by incrementally adding noise, and converging to a unit Gaussian distribution $N(\mathbf{0},\mathbf{I})$. New samples can be generated by reversing the forward diffusion process, by going from Gaussian noise back to the space of mask frames. To solve the reverse diffusion process, we need to train a noise predictor $\epsilon_\theta(\cdot |t)$ which is a time-conditioned U-net trained to predict the noise at each step of the diffusion process. The input to the network at step $t$ of the diffusion process is a channel-wise concatenation of the conditioning frames and noisy mask frames $\mathbf{[\mathbf{X_0,X_g},M_{1:K}^t]}$, and the output is the predicted noise of same dimensionality as the input. The training objective is as follows:
\begin{align*}
&\mathbb{E}_{t,\mathbf{[\mathbf{X_0,X_g},M_{1:K}]}\sim p_{\text{train}},\mathbf{\epsilon}\sim\mathcal{N(\mathbf{0},\mathbf{I})}}
\left[||\epsilon - \epsilon_\theta(\sqrt{\overline{\alpha}_t}\mathbf{M_{1:K}} + \sqrt{1-\overline{\alpha}_t}\mathbf{\epsilon} | \mathbf{X_0,X_g},t)||^2 \right]
\end{align*}

Here $\alpha$ is a constant hyper-parameter that depends on the noise schedule of the diffusion process.  The architecture of the U-Net for the Diffusion model is based on prior works~\cite{voleti2022masked,rombach2021highresolution}, and it uses a combination of 2D convolutions, multi-head self-attention layers, and adaptive group-norm. The noise levels ($p\in[0,1]$) use positional encodings that are adapted to the correct dimensionality for each residual block through fully connected layers.  The individual residual blocks in the U-Net consist of GroupNorm, conv layers, fully connected layers, and dropout, and follow the architecture in~\cite{voleti2022masked}.

For training the prediction model we obtain 2 second video clips from EpicKitchens~\cite{epic} and Ego4D~\cite{ego4d}. To obtain ground-truth hand-object masks, we use Visor annotations~\cite{visor} for EpicKitchens and an off-the-shelf predictor~\cite{hos} for obtaining the masks from Ego4D videos. In total, we curate around 150,000 video clips for training. The prediction model takes about 70 hours to train for 250,000 iterations on 8 2080Ti GPUs with a batch size of 64, and learning rate 1e-5.

\subsubsection{The Translation model}
The translation model is a transformer that is conditioned on the outputs of the future prediction model $M_{1:K}$ and for each observation $O_t$, predicts actions for $H$ steps in the future. The model behaves as a closed-loop policy that is queried at each time-step $t$ during deployment. The horizon lengths for each trajectory is 40, and we predict for $H=10$ horizon at each time-step. The observations are of resolution 224x224, and we process them with ResNet18 backbones to obtain features. We upsample the predicted masks from 64x64 to to 224x224 dimension images and process them also with ResNet18 CNNs. At each time-step we feed in a history of 3 steps, i.e. the past two observations and actions, and the current observation. The actions are of dimension 8 (7 for joint positions, and the 8th dimension for end-effector open/close). We directly predict target joint positions instead of delta positions, as shown to be helpful by recent work~\cite{act}.  The transformer encoder has 4 self-attention blocks, and the decoder has 7 cross-attention blocks, and the hidden dimensions are of size 512. We use a learning rate of 1e-5, batch size of 32, and dropout 0.1. 

\subsection{Baselines and Ablations}
We consider a goal-conditioned behavior cloning baseline that is not conditioned on the predicted masks, and is directly trained on all the robot data collected in-lab ($\sim$1400 trajectories). For the in-the-wild experiments, we additionally fine-tune the model with the 200 paired trajectories collected for these experiments. The architecture of the policy is a transformer similar to our translation model without the conditioning on hand-object masks, and keeping everything else the same. 

We consider another baseline that uses paired in-lab human-robot data, to be an adaptation of MimicPlay~\cite{wang2023mimicplay} . We train the latent planner model of MimicPlay (MP) with the human-data in the paired data of 400 trajectories we have collected for the experiments. For the in-the-wild experiments, we additionally fine-tune the model with the 200 paired trajectories collected for these experiments. Note that in the original paper~\cite{wang2023mimicplay}, there are a limited number of tasks (14) and human hand data is collected for 10 minutes per scene. In comparison, our paired data of 400 trajectories is much smaller and encompass around 40 tasks, since we focus mostly on learning from out-of-domain passive human videos from the web. We cannot use this large passive data for MimicPlay baseline as their framework relies on having the human videos in the exact same setup as the robot teleop data.  

We compare with two baselines that use passive human videos in different ways. The first comparison is with VRB~\cite{bahl2023affordances} by using the affordance model from the paper to do affordance conditioned imitation learning. The second comparison is a baseline that is trained entirely with passive human videos, for coarse manipulation (H2R)~\cite{bharadhwaj2023zero}.

In addition to these, for the table-top experiments we consider variations of our translation model trained on only paired human-robot data ($\sim$400 trajectories), only hallucinated data ($\sim$1000 trajectories), and combined paired and hallucinated data ($\sim$1400 trajectories). These ablations are on the same translation model architecture, and use manually specified hand trajectories transformed to hand-object masks through~\cite{hos}. We manually provide masks instead of the predictions from the human plan prediction model, in order to evaluate the translation model in isolation independent from the prediction model.

\subsection{Table-Top Robot Experiment Setup Details}
For the robot experiments, we use several everyday objects like doors, microwaves, bowls, spatulas, boxes, french presses etc. (Fig.~\ref{fig:tasks} has the overall list of objects), a fixed Intel Realsense camera in the scene, and a Franka Emika Panda arm operated through joint position control. We do not impose any artificial constraints on the robot's motions beyond what is possible without reaching joint limits. The action space of the translation model is 8 dimensional (7 for joint controls, and the 8th dimension for open/close of the gripper) We attach a Robotiq gripper to the arm with two festo finger grippers (for flexible grasps), so the overall end-effector is a two-finger gripper. As is the convention with image goals in real-robot experiments, we evaluate success by manually inspecting proximity of the final object configuration after robot execution, with that in the corresponding goal image. 

\subsection{In-The-Wild Robot Experiment Setup Details}
We use the same Franka Emika Panda arm with flexible two finger grippers as the previous table-top experiments. The only difference is that the robot is now mounted on a mobile base with four wheels that can be moved around. The same Intel Realsense camera is mounted next to the robot on the mobile base. We drag the robot across different kitchen and office scenes and perform experiments with the same setup described previously. Importantly, we do not modify the scenes and directly test on existing office and kitchen scenes. Please refer to the evaluation videos on the website for the diversity of manipulation skills and behaviors we are able to demonstrate with our framework. \href{https://homangab.github.io/hopman/}{https://homangab.github.io/hopman/}

\end{document}